\pdfoutput=1

\documentclass[11pt,a4paper]{article}

\usepackage[hyperref]{acl}
\usepackage{amsmath}
\DeclareMathOperator\arcosh{arcosh}
\usepackage{booktabs}
\usepackage{tabularx}
\newcolumntype{Y}{>{\centering\arraybackslash}X}
\usepackage{amssymb}
\usepackage{pifont}
\newcommand{\cmark}{\ding{51}}%
\newcommand{\xmark}{\ding{55}}%
\usepackage{times}
\usepackage{latexsym}

\usepackage[T1]{fontenc}

\usepackage[utf8]{inputenc}

\usepackage{microtype}

\title{Cross-lingual Word Embeddings in Hyperbolic Space}

\author{Chandni Saxena \\
  The Chinese University of Hong Kong\\
  \texttt{csaxena@cse.cuhk.edu.hk}
  \And
  Mudit Chaudhary \thanks{~~Work done during an assistantship at CUHK}\\
  University of Massachusetts Amherst\\
  \texttt{mchaudhary@umass.edu} 
  \AND
  Helen Meng\\
  The Chinese University of Hong Kong\\
  \texttt{hmmeng@se.cuhk.edu.hk} 
  }

\begin{document}
\maketitle
\begin{abstract}
Cross-lingual word embeddings can be applied to several natural language processing applications
across multiple languages. Unlike prior works that use word embeddings based on the Euclidean space, this short paper presents a simple and effective cross-lingual Word2Vec model that adapts to the Poincar{\'e} ball model of hyperbolic space to learn unsupervised cross-lingual word representations from a German-English parallel corpus. It has been shown that hyperbolic embeddings can capture and preserve hierarchical relationships.
We evaluate the model on both hypernymy and analogy tasks. 
The proposed model achieves comparable performance with the 
vanilla Word2Vec model on the cross-lingual analogy task, the hypernymy task shows that the cross-lingual Poincar{\'e}  Word2Vec model can capture latent hierarchical structure from free text across languages, which are absent from the Euclidean-based Word2Vec representations. Our results show that by preserving the latent hierarchical information, hyperbolic spaces can offer better representations for cross-lingual embeddings.
\end{abstract}
\section{Introduction}
In Natural Language Processing (NLP), cross-lingual word embeddings refer to the representations of words from two or more languages in a joint feature space. Prior works have demonstrated the use of these continuous representations in a variety of NLP tasks such as information retrieval~\cite{zoph2016transfer}, semantic textual similarity~\cite{cer2017semeval}, knowledge transfer~\cite{gu2018universal}, lexical analysis~\cite{dong2018cross}, plagiarism detection~\cite{alzahrani2020identifying}, etc. across different languages.

Natural language data possesses latent tree-like hierarchies in linguistic ontologies (e.g., hypernyms, hyponyms)~\cite{dhingra2018embedding,astefanoaei2020hyperbolic} such as the taxonomy of WordNet \cite{miller1998wordnet} for a language. 
From the statistics of word co-occurrence in training text, word embeddings models in Euclidean space can capture associations of words and their semantic relatedness. However, they fail to capture asymmetric word relations, including the latent hierarchical structure of words such as \textit{specificity} \cite{dhingra2018embedding}. For example, `bulldog' is more specific than `dog.'   
The use of non-Euclidean spaces has recently been advocated as alternatives to the conventional Euclidean space to infer latent hierarchy 
from the language data~\cite{nickel2017poincare,nickel2018learning,dhingra2018embedding,tifrea2018poincare}. Learning cross-lingual hierarchies such as cross-lingual
types-sub types and hypernyms-hyponyms,
is useful for tasks like cross-lingual lexical entailment, textual entailment, machine translation, etc. \cite{vulic-etal-2019-multilingual}.


This paper builds upon previous work in monolingual hyperbolic Word2Vec\footnote{The hyperbolic Word2Vec model is not described in \citet{tifrea2018poincare}'s paper, but available in the corresponding codebase} modeling from \citet{tifrea2018poincare} by learning cross-lingual hyperbolic embeddings from a parallel corpus, As a first step, we adopt the German-English parallel corpus from \citet{WOLK2014126}. We summarize the main contributions as follows: (1) To the best of our knowledge, we are the first to attempt at learning cross-lingual embeddings of natural language data using non-Euclidean geometry; (2)  we evaluate the hyperbolic embeddings on cross-lingual HyperLex hypernym task to evaluate its performance in learning latent hierarchies from free text and how a word's specificity correlates to its embedding's norm. We also compare the hyperbolic Word2Vec embeddings with the vanilla Word2Vec embeddings in the cross lingual analogy task.  All code\footnote{\url{https://github.com/muditchaudhary/hyperbolic_crosslingual_word_embeddings.git}} used are publicly available.

\section{Related Work}
 
 \subsection{Cross-lingual Word Embeddings} Cross-lingual word representations have been a subject of extensive research \cite{upadhyay2016cross,ruder2019survey}. Recent advances in the field can be grouped into unsupervised, supervised, and joint learning algorithms. Unsupervised models \cite{lample2017unsupervised,artetxe2019margin,chen2018adversarial} exploit existing monolingual word embeddings, followed by various cross-lingual alignment procedures.
 Supervised models \cite{mikolov2013exploiting,smith2017offline,grave2018learning} learn a mapping function from a source embedding space to the target embedding space based on different objective criteria. Joint learning models \cite{coulmance2015trans,josifoski2019crosslingual,sabet2019robust,lachraf2019arbengvec}  use parallel corpora to train bilingual embeddings in the same space jointly. This work adopts the settings similar to the joint learning model for embedding alignments by \citet{lachraf2019arbengvec}. 

 \subsection{Hyperbolic Word Embeddings} Hyperbolic spaces offer a continuous representation for embedding tree-like structures with arbitrarily low distortion~\cite{sala2018representation,chami2020trees}. Word embeddings in hyperbolic spaces have been applied to diverse NLP applications such as text classification~\cite{zhu2020hypertext}, learning taxonomy~\cite{astefanoaei2020hyperbolic}, and concept hierarchy~\cite{le2019inferring}. By using hyperbolic space these applications were able to outperform their euclidean counterparts by exploiting the benefits of hierarchical structure of the text data with high quality embedding which capture similarity and generality of concept together enforce transitivity of the is-a-relations in a smaller embedding space \cite{le2019inferring}. 
 Some recent work use supervised models \cite{nickel2017poincare,nickel2018learning,ganea2018hyperbolic} that require external information on word relations such as WordNet or ConceptNet in addition to free text corpora to learn word and sentence embeddings in the hyperbolic space. 
 \citet{nickel2017poincare} consider a non-parametric method to learn hierarchical representation from a lookup table for symbolic data. \citet{ganea2018hyperbolic} propose a supervised method to learn embeddings for an acyclic graph structure of words. Unsupervised word embedding models \cite{leimeister2018skip,dhingra2018embedding,tifrea2018poincare} which can directly learn from text corpora
 have been recently applied in the hyperbolic spaces. \citet{leimeister2018skip} employ the skip-gram with negative sampling architecture of the Word2Vec model for learning word embeddings from free text. \citet{dhingra2018embedding} present a two-step model to embed a co-occurrence graph of words and map the output of the encoder to the Poincar{\'e} ball using the algorithm from~\citet{nickel2017poincare}. \citet{tifrea2018poincare} remodel the GloVe algorithm to learn unsupervised word representation in hyperbolic spaces.
\section{Methodology}
\subsection{Hyperbolic Space}
Hyperbolic space in Riemannian geometry is a homogeneous space of constant negative curvature with special geometric properties. Hyperbolic space can endow infinite trees to have nearly isometric embeddings. We embed words using the Poincar{\'e} ball model of the hyperbolic space.\\
\textbf{The Poincar{\'e} Ball.} The Poincar{\'e} ball model $\mathcal{B}^n$ of $n$-dimensional hyperbolic geometry is a manifold equipped with a Riemannian metric $g^B$. Formally, an $n$-dimensional Poincar{\'e} unit ball is defined as $(\mathcal{B}^n, g^B)$ and the metric $g^B$ is conformal to the Euclidean metric $g^E$ as $g^B= {\lambda_x}^2. g^E$. Where $\lambda_x = \frac{2}{1-||x||^2}$, $x \in \mathcal{B}^n$, and  $||.||$ stands for the Euclidean norm.
Notably, the hyperbolic distance $d_{\mathcal{B}^n}$ between $n$-dimensional points $(x, y) \in\mathcal{B}^n $ in the Poincaré ball is defined as:
    \begin{equation} \small \label{eq:1}
    d_{\mathcal{B}^n}(x,y) = \arcosh \left ( 1+2\frac{||x-y||^2}{(1-||x||^2)(1-||y||^2)}\right )
    \end{equation}
 where $\arcosh (w)= \ln (w+\sqrt{w^2-1})$ is the inverse of hyperbolic cosine function. Using ambient Euclidean geometry, the geodesic distance between points $(x,y)$ can be induced using Equation~(\ref{eq:1}) as  $d_{\mathcal{B}^n}(x,y)= \arcosh \left (
 1+ \frac{1}{2}{\lambda_x}{\lambda_y||x-y||^2} \right )$. This indicates that the distance changes evenly w.r.t. $||x||$ and $||y||$, which is a key point to learning continuous representation for hierarchical structures~\cite{chen2020hyperbolic, saxena2020survey}. \\ 
\subsection{Hyperbolic Cross-lingual Word Embedding} \label{hyperbolic-methodology}
We first adopt the mono-lingual hyperbolic word embedding from a model defined in the work by \citet{tifrea2018poincare}. We extend it to cross-lingual hyperbolic word embedding by using parallel text corpora input to capture word relationsships through bilingual word co-occurrence statistics. \citet{tifrea2018poincare} added a hyperparameter function $h$ on the distance between word and context pairs in the hyperbolic Word2Vec's objective function. Hence, the effective distance function in the objective function becomes $h(d_{\mathcal{B}^n}(x,y))$.

Hyperbolic word embeddings have shown to embed general words near the origin and specific words towards the edges -- we attempt to exploit this property to identify latent hierarchies and in hypernym evaluation task by using the Poincar{\'e} norms of the words to determine their hierarchy as words with higher norm will be more specific, i.e., lower in hierarchy \cite{nickel2017poincare, dhingra2018embedding, linzhuo2020social}. We evaluate the hyperbolic model on the cross-lingual analogy task to compare it with its Euclidean counterpart.

\subsection{Cross-lingual Alignment}
To train the cross-lingual Word2Vec model in the hyperbolic space, we perform a pre-processing step of word-to-word alignment as defined by \citet{lachraf2019arbengvec} using parallel sentences from a bilingual parallel corpus. 
We generate word-to-word alignment by matching the indices of tokens from both languages in parallel sentences.

\subsection{Evaluation Methodology}
\textbf{Hypernymy Evaluation.}
We perform hypernymy evaluation to assess performance of the proposed model based on learning the latent hierarchical structure from free text.
In the hypernymy evaluation task, given a word pair $(u,v)$, we evaluate $is$-$a(u,v)$ i.e., to what degree $u$ is of type $v$. 

For English, German and cross-lingual German-English hypernymy evaluation, we use the HyperLex benchmark \citet{vulic-etal-2017-hyperlex, vulic-etal-2019-multilingual}, which contains word pairs $(u,v)$ and a corresponding degree to which $u$ is of type $v$ i.e. the $is$-$a$ score. This score has been obtained by human annotators, scored by the degree of typicality and semantic category membership \cite{vulic-etal-2017-hyperlex}. For example, in the HyperLex dataset, $is$-$a(chemistry,science)=6.00$ and $is$-$a(chemistry, knife)=0.50$ as chemistry is a type of science but not a type of knife. 

To generate the $is$-$a$ score we follow the same approach as used by \citet{nickel2017poincare}:
\begin{equation}
    \textit{$is$-$a$}(u,v)= -(1+\alpha(||v||-||u||))d_{\mathcal{B}^n}(u,v)
\end{equation}

The evaluation is performed by calculating the Spearman correlation between the ground-truth score and the predicted score. Note that our model is not trained on any hypernymy detection task but tries to learn latent hierarchy from free text. 

\textbf{Cross-lingual Analogy Evaluation.}
The analogy evaluation task is one of the standard intrinsic evaluations for word embeddings. In cross-lingual analogy evaluation task, given a word pair $(w_1,w_2)$ in one language, and a word $w_3$ in the other language, the goal is to predict the word $w_4^*$ such that $w_4^*$ is related to $w_3$ same way $w_2$ is related to $w_1$. For example, as prince ($w_1$) is to princess ($w_2$), prinz ($w_3$; German equivalent for \textit{prince}) is to prinzessin ($w_4^*$; German equivalent for \textit{princess}). For evaluating cross-lingual analogy for the German and English language, we use the cross-lingual analogy dataset provided by \citet{brychcin2018crosslingual}. 

\section{Experiments \& Results}
\begin{table}
\small
\begin{tabularx}{\columnwidth}{c|X}
\toprule
\textbf{Word} & \textbf{Closest Children} \\ \midrule
Species  &       arten, gattung, subspecies, unterfamilie \\
Physics  &       astrophysik, astrophysics, mechanik \\ 
Molekülen    &       atomen, protonen, elektronen, ionen  \\
Orchestra  &      symphony, philharmonic, concerto \\ 
Regierung    &       governments, regierungen, bundesregierung \\ \bottomrule
\end{tabularx}
\caption{\label{closest-children}
For a given word in the left column, this table shows the top closest children using a 100Dim with bias hyperbolic Word2Vec model. Note that the children consist of both English and German words.}
\end{table}
\begin{table}
\centering
\begin{tabularx}{\columnwidth}{lYYY}
\toprule
& \multicolumn{3}{c}{\textbf{HyperLex}} \\
\textbf{Hyperbolic Model}   & {\textbf{English}} en & \textbf{German} de& \textbf{Cross} de-en \\
\midrule
100D & 0.166 & 0.130& 0.150 \\
100D w/ bias  & 0.175 & 0.104& 0.162  \\
120D w/ bias & \textbf{0.192} & 0.120& \textbf{0.179}\\
300D w/ bias &0.183 &\textbf{0.125} & 0.155  \\ 
\bottomrule
\end{tabularx}

\caption{\label{hypernymy-results}
Spearman correlations from different hyperbolic Word2Vec models on the English, German and German-English HyperLex dataset for hypernymy evaluation. Best results are in bold.}
\end{table}
\begin{table*}[ht]
\small

\begin{tabularx}{\textwidth}{YYYYYYYYYYYY}
\toprule
   \multicolumn{3}{c}{\textbf{``music''}}  & \multicolumn{3}{c}{\textbf{``art''}}  & \multicolumn{3}{c}{\textbf{``film''}} & \multicolumn{3}{c}{\textbf{``chemistry''}}\\ \cmidrule(r){1-3} \cmidrule(l){4-6} \cmidrule(l){7-9} \cmidrule(l){10-12}
    \textbf{Word} & \textbf{Count} & \textbf{Norm}           & \textbf{Word} & \textbf{Count} & \textbf{Norm}          & \textbf{Word} & \textbf{Count} & \textbf{Norm}          & \textbf{Word} & \textbf{Count} & \textbf{Norm}   \\ \cmidrule(r){1-3} \cmidrule(l){4-6} \cmidrule(l){7-9} \cmidrule(l){10-12} 
    
    music & 33167 & 0.607 & art & 28551 & 0.606  & film & 61682 & 0.606 & chemistry & 3165 & 0.628 \\
    musik & 10637 & 0.608 & arts & 13888 & 0.623 & films & 7185 & 0.607 & chemie & 2530 & 0.629 \\
    musical & 6585 & 0.612 & design & 11558 & 0.624 & drama & 4948 & 0.617 & chemiker & 908 & 0.620\\ 
    musicians & 1955 & 0.628 & skulptur & 480 & 0.632 & comedy & 3937 & 0.630 & chemischen & 628 & 0.647\\
    filmmusik & 278 & 0.640 & kunstgalerie & 102 & 0.665 & stummfilm & 179 & 0.648 & organischen & 344 & 0.651\\

\bottomrule
\end{tabularx}

\caption{\label{specificit-results}
 Words in order of increasing hyperbolic norm which are related to the word indicated in the top row along with their counts in the corpus. General words have a lower norm and specific words have a higher norm.}
\end{table*}

\begin{table}
\centering
\begin{tabularx}{\columnwidth}{llYY}
\toprule
\textbf{Model Type} & \textbf{Dim} & \textbf{Bias term} & {\textbf{Accuracy}}\\
\hline
Vanilla & 20D & \xmark & 16.8\\ 
Poincaré & 20D & \xmark & 20.5\\
Vanilla & 40D & \xmark & 25.4\\ 
Poincaré & 40D & \xmark & 26.5\\
Vanilla & 80D & \xmark & 30.8\\ 
Poincaré & 80D & \xmark & 28.7\\
Vanilla & 180D & \cmark & 36.1\\ 
Poincaré & 180D & \cmark & 29.3\\
\bottomrule
\end{tabularx}
\caption{\label{analogy-results}
Accuracy on the cross-lingual analogy task.}
\end{table}

\subsection{Dataset}

This paper uses the Wikipedia corpus of parallel sentences extracted by \citet{WOLK2014126} to train the model. The dataset is accessed through OPUS \cite{TIEDEMANN12.463}. The corpus consists of \textasciitilde2.5 million parallel aligned German-English sentence pairs with 43.5 million German tokens and 58.4 million English tokens.

\subsection{Experimental Settings}
We reference \citet{tifrea2018poincare}'s Poincar{\'e} Word2Vec implementation\footnote{https://github.com/alex-tifrea/poincare\_glove} and extended it to learn cross-lingual word embeddings.  
We set the minimum frequency of words in the vocabulary to 100, and a window size of 5. The models use Negative-Log-Likelihood loss. The non-hyperbolic vanilla Word2Vec uses Stochastic Gradient Descent optimizer, whereas hyperbolic Word2Vec uses Weighted Full Riemannian Stochastic Gradient Descent optimizer \cite{Bonnabel_2013}. For hyperbolic embeddings, the hyperparameter $h$ is set to $cosh^2(x)$. 
During the analogy evaluation, we use the cosine distance instead of Poincar{\'e} distance for hyperbolic models. 
We use the hypernymysuite\footnote{https://github.com/facebookresearch/hypernymysuite} for hypernymy evaluation \cite{roller2018hearst}.

\subsection{Evaluation Results}
\textbf{Hypernymy Evaluation.}
We present the top closest children of selected words in Table \ref{closest-children}. As described in Section \ref{hyperbolic-methodology}, the closest children are calculated by finding the target word's $(t)$ nearest neighbours $(N)$ and extracting the neighbour $n \in N$ such that $||n||_p > ||t||_p$, where $||.||_p$ is the Poincar{\'e} norm. We observe that the model is able to find the hyponyms of the words using the closest children across languages. For example, the children of `Physics' are its subtypes -- `astrophysik' (astrophysics), `astrophysics', `mechanik' (mechanics), and `biophysics'.

Table \ref{hypernymy-results} reports the results on the hypernymy evaluation task. Although the models were not trained on hypernymy tasks, we observe that they could still learn some latent hierarchies from the free text across languages. Word pairs with out-of-vocabulary words were ignored during evaluation.

Table \ref{specificit-results} shows lists of related words in order of increasing hyperbolic norm and specificity, similar to \citet{dhingra2018embedding}'s evaluation. We show counts of these words in the corpus. Higher the count, more generic the word, and has a smaller hyperbolic norm. The Spearman correlation between 1/$f$, where $f$ is the frequency of a word in the corpus, and its embedding's hyperbolic norm is $0.747$ using a 300D w/bias Poincaré model. 

\textbf{Cross-lingual Analogy Evaluation.} Table \ref{analogy-results} reports the results on the cross-lingual analogy task. We observe that for 20D models, hyperbolic model outperformed the vanilla model. For higher dimension models, hyperbolic Word2Vec performed on par with its Euclidean counterpart. Similar to hypernymy evaluation, analogy pairs with out-of-vocabulary words were ignored during evaluation.

\section{Conclusion and Future Work}


This work adapts a monolingual hyperbolic Word2Vec model and extend to cross-lingual embeddings.
We observe that the hyperbolic Word2Vec embeddings are competent on cross-lingual analogy task. The hypernymy evaluation show that it also captures some latent hierarchies across languages without being trained on a hypernymy task. Future work will include extrinsic evaluation of hyperbolic cross-lingual word embeddings on downstream tasks such as machine translation, cross-lingual textual entailment detection, cross-lingual taxonomy learning, etc.


\begin{thebibliography}{39}
\expandafter\ifx\csname natexlab\endcsname\relax\def\natexlab#1{#1}\fi

\bibitem[{Alzahrani and Aljuaid(2020)}]{alzahrani2020identifying}
Salha Alzahrani and Hanan Aljuaid. 2020.
\newblock Identifying cross-lingual plagiarism using rich semantic features and
  deep neural networks: A study on arabic-english plagiarism cases.
\newblock \emph{Journal of King Saud University-Computer and Information
  Sciences}.

\bibitem[{Artetxe and Schwenk(2019)}]{artetxe2019margin}
Mikel Artetxe and Holger Schwenk. 2019.
\newblock Margin-based parallel corpus mining with multilingual sentence
  embeddings.
\newblock In \emph{Proceedings of the 57th Annual Meeting of the Association
  for Computational Linguistics}, pages 3197--3203.

\bibitem[{Astefanoaei and Collignon(2020)}]{astefanoaei2020hyperbolic}
Maria Astefanoaei and Nicolas Collignon. 2020.
\newblock Hyperbolic embeddings for music taxonomy.
\newblock In \emph{Proceedings of the 1st Workshop on NLP for Music and Audio
  (NLP4MusA)}, pages 38--42.

\bibitem[{Bonnabel(2013)}]{Bonnabel_2013}
Silvere Bonnabel. 2013.
\newblock \href {https://doi.org/10.1109/tac.2013.2254619} {Stochastic gradient
  descent on riemannian manifolds}.
\newblock \emph{IEEE Transactions on Automatic Control}, 58(9):2217–2229.

\bibitem[{Brychcín et~al.(2018)Brychcín, Taylor, and
  Svoboda}]{brychcin2018crosslingual}
Tomáš Brychcín, Stephen~Eugene Taylor, and Lukáš Svoboda. 2018.
\newblock \href {http://arxiv.org/abs/1807.04175} {Cross-lingual word analogies
  using linear transformations between semantic spaces}.

\bibitem[{Cer et~al.(2017)Cer, Diab, Agirre, Lopez-Gazpio, and
  Specia}]{cer2017semeval}
Daniel Cer, Mona Diab, Eneko Agirre, I{\~n}igo Lopez-Gazpio, and Lucia Specia.
  2017.
\newblock Semeval-2017 task 1: Semantic textual similarity multilingual and
  crosslingual focused evaluation.
\newblock In \emph{Proceedings of the 11th International Workshop on Semantic
  Evaluation (SemEval-2017)}, pages 1--14.

\bibitem[{Chami et~al.(2020)Chami, Gu, Chatziafratis, and
  R{\'e}}]{chami2020trees}
Ines Chami, Albert Gu, Vaggos Chatziafratis, and Christopher R{\'e}. 2020.
\newblock From trees to continuous embeddings and back: Hyperbolic hierarchical
  clustering.
\newblock \emph{arXiv preprint arXiv:2010.00402}.

\bibitem[{Chen et~al.(2020)Chen, Huang, Xiao, Cai, and
  Jing}]{chen2020hyperbolic}
Boli Chen, Xin Huang, Lin Xiao, Zixin Cai, and Liping Jing. 2020.
\newblock Hyperbolic interaction model for hierarchical multi-label
  classification.
\newblock In \emph{Proceedings of the AAAI Conference on Artificial
  Intelligence}, volume~34, pages 7496--7503.

\bibitem[{Chen et~al.(2018)Chen, Sun, Athiwaratkun, Cardie, and
  Weinberger}]{chen2018adversarial}
Xilun Chen, Yu~Sun, Ben Athiwaratkun, Claire Cardie, and Kilian Weinberger.
  2018.
\newblock Adversarial deep averaging networks for cross-lingual sentiment
  classification.
\newblock \emph{Transactions of the Association for Computational Linguistics},
  6:557--570.

\bibitem[{Coulmance et~al.(2015)Coulmance, Marty, Wenzek, and
  Benhalloum}]{coulmance2015trans}
Jocelyn Coulmance, Jean-Marc Marty, Guillaume Wenzek, and Amine Benhalloum.
  2015.
\newblock Trans-gram, fast cross-lingual word-embeddings.
\newblock In \emph{Proceedings of the 2015 Conference on Empirical Methods in
  Natural Language Processing}, pages 1109--1113.

\bibitem[{Dhingra et~al.(2018)Dhingra, Shallue, Norouzi, Dai, and
  Dahl}]{dhingra2018embedding}
Bhuwan Dhingra, Christopher Shallue, Mohammad Norouzi, Andrew Dai, and George
  Dahl. 2018.
\newblock Embedding text in hyperbolic spaces.
\newblock In \emph{Proceedings of the Twelfth Workshop on Graph-Based Methods
  for Natural Language Processing (TextGraphs-12)}, pages 59--69.

\bibitem[{Dong and De~Melo(2018)}]{dong2018cross}
Xin Dong and Gerard De~Melo. 2018.
\newblock Cross-lingual propagation for deep sentiment analysis.
\newblock In \emph{Proceedings of the AAAI Conference on Artificial
  Intelligence}, volume~32.

\bibitem[{Ganea et~al.(2018)Ganea, B{\'e}cigneul, and
  Hofmann}]{ganea2018hyperbolic}
Octavian Ganea, Gary B{\'e}cigneul, and Thomas Hofmann. 2018.
\newblock Hyperbolic entailment cones for learning hierarchical embeddings.
\newblock In \emph{International Conference on Machine Learning}, pages
  1646--1655. PMLR.

\bibitem[{Grave et~al.(2018)Grave, Bojanowski, Gupta, Joulin, and
  Mikolov}]{grave2018learning}
{\'E}douard Grave, Piotr Bojanowski, Prakhar Gupta, Armand Joulin, and
  Tom{\'a}{\v{s}} Mikolov. 2018.
\newblock Learning word vectors for 157 languages.
\newblock In \emph{Proceedings of the Eleventh International Conference on
  Language Resources and Evaluation (LREC 2018)}.

\bibitem[{Gu et~al.(2018)Gu, Hassan, Devlin, and Li}]{gu2018universal}
Jiatao Gu, Hany Hassan, Jacob Devlin, and Victor~OK Li. 2018.
\newblock Universal neural machine translation for extremely low resource
  languages.
\newblock In \emph{Proceedings of the 2018 Conference of the North American
  Chapter of the Association for Computational Linguistics: Human Language
  Technologies, Volume 1 (Long Papers)}, pages 344--354.

\bibitem[{Josifoski et~al.(2019)Josifoski, Paskov, Paskov, Jaggi, and
  West}]{josifoski2019crosslingual}
Martin Josifoski, Ivan~S Paskov, Hristo~S Paskov, Martin Jaggi, and Robert
  West. 2019.
\newblock Crosslingual document embedding as reduced-rank ridge regression.
\newblock In \emph{Proceedings of the Twelfth ACM International Conference on
  Web Search and Data Mining}, pages 744--752.

\bibitem[{Lachraf et~al.(2019)Lachraf, Ayachi, Abdelali, Schwab
  et~al.}]{lachraf2019arbengvec}
Raki Lachraf, Youcef Ayachi, Ahmed Abdelali, Didier Schwab, et~al. 2019.
\newblock Arbengvec: Arabic-english cross-lingual word embedding model.
\newblock In \emph{Proceedings of the Fourth Arabic Natural Language Processing
  Workshop}, pages 40--48.

\bibitem[{Lample et~al.(2017)Lample, Conneau, Denoyer, and
  Ranzato}]{lample2017unsupervised}
Guillaume Lample, Alexis Conneau, Ludovic Denoyer, and Marc'Aurelio Ranzato.
  2017.
\newblock Unsupervised machine translation using monolingual corpora only.
\newblock \emph{arXiv preprint arXiv:1711.00043}.

\bibitem[{Le et~al.(2019)Le, Roller, Papaxanthos, Kiela, and
  Nickel}]{le2019inferring}
Matthew Le, Stephen Roller, Laetitia Papaxanthos, Douwe Kiela, and Maximilian
  Nickel. 2019.
\newblock Inferring concept hierarchies from text corpora via hyperbolic
  embeddings.
\newblock In \emph{Proceedings of the 57th Annual Meeting of the Association
  for Computational Linguistics}, pages 3231--3241.

\bibitem[{Leimeister and Wilson(2018)}]{leimeister2018skip}
Matthias Leimeister and Benjamin~J Wilson. 2018.
\newblock Skip-gram word embeddings in hyperbolic space.
\newblock \emph{arXiv preprint arXiv:1809.01498}.

\bibitem[{Linzhuo et~al.(2020)Linzhuo, Lingfei, and James}]{linzhuo2020social}
Li~Linzhuo, Wu~Lingfei, and Evans James. 2020.
\newblock Social centralization and semantic collapse: Hyperbolic embeddings of
  networks and text.
\newblock \emph{Poetics}, 78:101428.

\bibitem[{Mikolov et~al.(2013)Mikolov, Le, and
  Sutskever}]{mikolov2013exploiting}
Tomas Mikolov, Quoc~V Le, and Ilya Sutskever. 2013.
\newblock Exploiting similarities among languages for machine translation.
\newblock \emph{arXiv preprint arXiv:1309.4168}.

\bibitem[{Miller(1998)}]{miller1998wordnet}
George~A Miller. 1998.
\newblock \emph{WordNet: An electronic lexical database}.
\newblock MIT press.

\bibitem[{Nickel and Kiela(2017)}]{nickel2017poincare}
Maximillian Nickel and Douwe Kiela. 2017.
\newblock Poincar{\'e} embeddings for learning hierarchical representations.
\newblock \emph{Advances in Neural Information Processing Systems},
  30:6338--6347.

\bibitem[{Nickel and Kiela(2018)}]{nickel2018learning}
Maximillian Nickel and Douwe Kiela. 2018.
\newblock Learning continuous hierarchies in the lorentz model of hyperbolic
  geometry.
\newblock In \emph{International Conference on Machine Learning}, pages
  3779--3788. PMLR.

\bibitem[{Roller et~al.(2018)Roller, Kiela, and Nickel}]{roller2018hearst}
Stephen Roller, Douwe Kiela, and Maximilian Nickel. 2018.
\newblock Hearst patterns revisited: Automatic hypernym detection from large
  text corpora.
\newblock In \emph{Proceedings of the 56th Annual Meeting of the Association
  for Computational Linguistics}. Association for Computational Linguistics.

\bibitem[{Ruder et~al.(2019)Ruder, Vuli{\'c}, and S{\o}gaard}]{ruder2019survey}
Sebastian Ruder, Ivan Vuli{\'c}, and Anders S{\o}gaard. 2019.
\newblock A survey of cross-lingual word embedding models.
\newblock \emph{Journal of Artificial Intelligence Research}, 65:569--631.

\bibitem[{Sabet et~al.(2019)Sabet, Gupta, Cordonnier, West, and
  Jaggi}]{sabet2019robust}
Ali Sabet, Prakhar Gupta, Jean-Baptiste Cordonnier, Robert West, and Martin
  Jaggi. 2019.
\newblock Robust cross-lingual embeddings from parallel sentences.
\newblock \emph{arXiv preprint arXiv:1912.12481}.

\bibitem[{Sala et~al.(2018)Sala, De~Sa, Gu, and
  R{\'e}}]{sala2018representation}
Frederic Sala, Chris De~Sa, Albert Gu, and Christopher R{\'e}. 2018.
\newblock Representation tradeoffs for hyperbolic embeddings.
\newblock In \emph{International conference on machine learning}, pages
  4460--4469. PMLR.

\bibitem[{Saxena et~al.(2020)Saxena, Liu, and King}]{saxena2020survey}
Chandni Saxena, Tianyu Liu, and Irwin King. 2020.
\newblock A survey of graph curvature and embedding in non-euclidean spaces.
\newblock In \emph{International Conference on Neural Information Processing},
  pages 127--139. Springer.

\bibitem[{Smith et~al.(2017)Smith, Turban, Hamblin, and
  Hammerla}]{smith2017offline}
Samuel~L Smith, David~HP Turban, Steven Hamblin, and Nils~Y Hammerla. 2017.
\newblock Offline bilingual word vectors, orthogonal transformations and the
  inverted softmax.
\newblock \emph{arXiv preprint arXiv:1702.03859}.

\bibitem[{Tiedemann(2012)}]{TIEDEMANN12.463}
Jörg Tiedemann. 2012.
\newblock Parallel data, tools and interfaces in opus.
\newblock In \emph{Proceedings of the Eight International Conference on
  Language Resources and Evaluation (LREC'12)}, Istanbul, Turkey. European
  Language Resources Association (ELRA).

\bibitem[{Tifrea et~al.(2018)Tifrea, Becigneul, and Ganea}]{tifrea2018poincare}
Alexandru Tifrea, Gary Becigneul, and Octavian-Eugen Ganea. 2018.
\newblock Poincare glove: Hyperbolic word embeddings.
\newblock In \emph{International Conference on Learning Representations}.

\bibitem[{Upadhyay et~al.(2016)Upadhyay, Faruqui, Dyer, and
  Roth}]{upadhyay2016cross}
Shyam Upadhyay, Manaal Faruqui, Chris Dyer, and Dan Roth. 2016.
\newblock Cross-lingual models of word embeddings: An empirical comparison.
\newblock In \emph{Proceedings of the 54th Annual Meeting of the Association
  for Computational Linguistics (Volume 1: Long Papers)}, pages 1661--1670.

\bibitem[{Vuli{\'c} et~al.(2017)Vuli{\'c}, Gerz, Kiela, Hill, and
  Korhonen}]{vulic-etal-2017-hyperlex}
Ivan Vuli{\'c}, Daniela Gerz, Douwe Kiela, Felix Hill, and Anna Korhonen. 2017.
\newblock \href {https://doi.org/10.1162/COLI_a_00301} {{H}yper{L}ex: A
  large-scale evaluation of graded lexical entailment}.
\newblock \emph{Computational Linguistics}, 43(4):781--835.

\bibitem[{Vuli{\'c} et~al.(2019)Vuli{\'c}, Ponzetto, and
  Glava{\v{s}}}]{vulic-etal-2019-multilingual}
Ivan Vuli{\'c}, Simone~Paolo Ponzetto, and Goran Glava{\v{s}}. 2019.
\newblock \href {https://doi.org/10.18653/v1/P19-1490} {Multilingual and
  cross-lingual graded lexical entailment}.
\newblock In \emph{Proceedings of the 57th Annual Meeting of the Association
  for Computational Linguistics}, pages 4963--4974, Florence, Italy.
  Association for Computational Linguistics.

\bibitem[{Wołk and Marasek(2014)}]{WOLK2014126}
Krzysztof Wołk and Krzysztof Marasek. 2014.
\newblock \href {https://doi.org/https://doi.org/10.1016/j.protcy.2014.11.024}
  {Building subject-aligned comparable corpora and mining it for truly parallel
  sentence pairs}.
\newblock \emph{Procedia Technology}, 18:126--132.
\newblock International workshop on Innovations in Information and
  Communication Science and Technology, IICST 2014, 3-5 September 2014, Warsaw,
  Poland.

\bibitem[{Zhu et~al.(2020)Zhu, Zhou, Xiao, Jiang, Chen, and
  Liu}]{zhu2020hypertext}
Yudong Zhu, Di~Zhou, Jinghui Xiao, Xin Jiang, Xiao Chen, and Qun Liu. 2020.
\newblock Hypertext: Endowing fasttext with hyperbolic geometry.
\newblock In \emph{Proceedings of the 2020 Conference on Empirical Methods in
  Natural Language Processing: Findings}, pages 1166--1171.

\bibitem[{Zoph et~al.(2016)Zoph, Yuret, May, and Knight}]{zoph2016transfer}
Barret Zoph, Deniz Yuret, Jonathan May, and Kevin Knight. 2016.
\newblock Transfer learning for low-resource neural machine translation.
\newblock In \emph{Proceedings of the 2016 Conference on Empirical Methods in
  Natural Language Processing}, pages 1568--1575.

\end{thebibliography}

\appendix



\end{document}